\documentclass[letterpaper]{article} % DO NOT CHANGE THIS
\usepackage{aaai2027}  % DO NOT CHANGE THIS
\nocopyright
\usepackage[hyphens]{url}  % DO NOT CHANGE THIS
\usepackage{graphicx} % DO NOT CHANGE THIS
\usepackage{natbib}  % DO NOT CHANGE THIS AND DO NOT ADD ANY OPTIONS TO IT
\usepackage{caption} % DO NOT CHANGE THIS AND DO NOT ADD ANY OPTIONS TO IT
\usepackage{amsmath}
\usepackage{amssymb}
\usepackage{booktabs}
\usepackage{multirow}
\usepackage{colortbl}
\usepackage{verbatim}
\usepackage{float}

\definecolor{DefectRow}{RGB}{255,242,242}
\definecolor{CleanRow}{RGB}{241,250,244}
\definecolor{GroupRow}{RGB}{242,242,242}
\definecolor{OursRow}{RGB}{232,244,255}
\definecolor{DGreen}{RGB}{23,140,77}
\definecolor{DRed}{RGB}{208,59,59}

\newcommand{\methodname}{NTEP-R}
\newcommand{\modelname}{NTEP-8B}
\newcommand{\toolcall}[1]{\texttt{#1}}

\title{Making Every Tool Call Count: Necessary Tool-Evidence Path Rewards for Agentic Vision-Language Models}
\author{
    Xingming Long\textsuperscript{\rm 3}\equalcontrib,
    Yu Liu\textsuperscript{\rm 1,3}\equalcontrib\corresponding,
    Zhiwei Yang\textsuperscript{\rm 1,3}\equalcontrib,
    Hanqi Feng\textsuperscript{\rm 2},
    Shaojie Zhang\textsuperscript{\rm 3}, \\
    Barnabas Poczos\textsuperscript{\rm 2},
    Chao Jiang\textsuperscript{\rm 3},
    Zhenbo Luo\textsuperscript{\rm 3},
    Lei Jiang\textsuperscript{\rm 1},
    Pei Fu\textsuperscript{\rm 3}\corresponding
}
\affiliations{
    Institute of Information Engineering, Chinese Academy of Sciences\textsuperscript{1}\\
    Department of Machine Learning, Carnegie Mellon University\textsuperscript{2}\\
    MiLM Plus, Xiaomi Inc.\textsuperscript{3}
}

\begin{document}

\maketitle

\begin{abstract}
Modern vision-language models (VLMs) can directly answer many image-grounded questions, yet they often struggle with complex queries requiring fine-grained visual details or external knowledge. To acquire this missing evidence, agentic VLMs invoke tools such as image cropping, image search, and text search. However, existing training paradigms primarily evaluate tool-use based on final answer correctness, leaving evidence acquisition and utilization insufficiently supervised. This leads to two critical shortcomings: \textbf{(i)} models frequently issue redundant or off-target tool calls that fail to gather necessary evidence, and \textbf{(ii)} even when appropriate tools are called, models often fail to extract the required information from the resulting observations.
To address these limitations, we introduce the \textbf{NTEP} (\textbf{N}ecessary \textbf{T}ool-\textbf{E}vidence \textbf{P}ath), a novel annotation scheme that explicitly specifies the essential external evidence and corresponding tool calls for each query. Building upon this, we propose \textbf{NTEP-R} (\textbf{NTEP} \textbf{R}eward), a supervision mechanism ensuring that each tool invocation strictly advances the reasoning process toward the final solution. Specifically, our approach rewards the agent for aligning its pre-call intent with a necessary evidence-seeking goal, and for ensuring the information summarized from the post-call observation aligns with the necessary evidence. Furthermore, we introduce a non-repeated-goal regularizer to penalize redundant calls that revisit satisfied NTEP goals.
Extensive evaluations on seven image-grounded benchmarks demonstrate that our 8B-parameter instantiation, \textbf{NTEP-8B}, significantly improves both search-oriented accuracy and tool-use efficiency within a unified three-tool framework. These results highlight the critical value of fine-grained tool-evidence path supervision for training robust agentic VLMs.
\end{abstract}
\begin{figure}[t]
\centering
\includegraphics[width=\columnwidth]{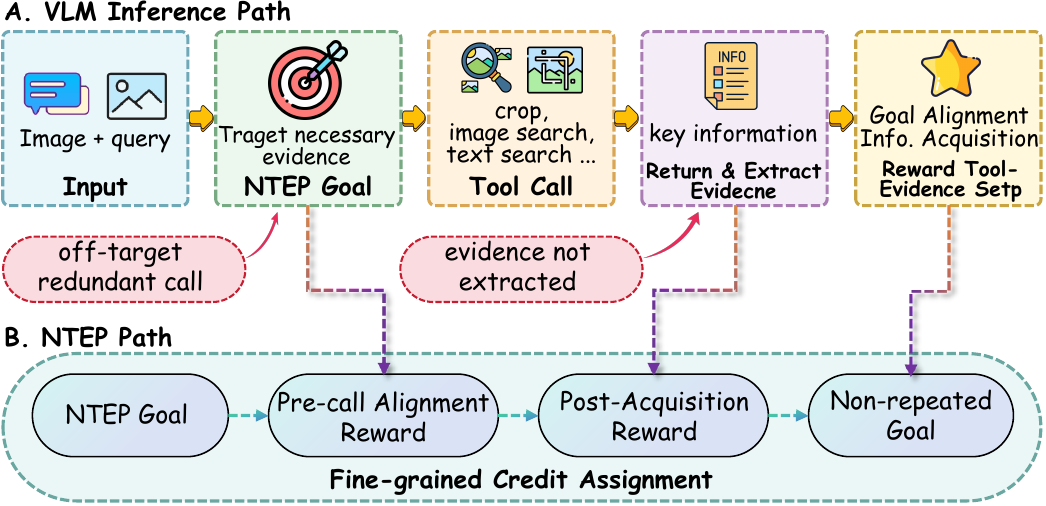}
\caption{Motivation for NTEP-R. A useful tool call must target necessary evidence before execution and extract the required information from the returned observation afterward; NTEP-R supervises both stages and penalizes repeated goals.}
\label{fig:ntepr_overview}
\end{figure}

\section{Introduction}
% 背景，引出为什么做agentic vlm（tool use）
Modern vision-language models (VLMs) demonstrate remarkable proficiency in answering image-grounded questions directly from initial image-query inputs~\cite{bai2025qwen3vl,comanici2025gemini25}. However, not all questions can be fully resolved relying solely on this provided context. Complex queries often hinge on fine-grained visual details requiring closer inspection, unfamiliar entities demanding identification, or external world knowledge necessitating verification. These limitations naturally motivate the development of agentic VLMs, which actively invoke tools to obtain external evidence during the reasoning process~\cite{wu2025mmsearchr1,chng2025sensenovamars,hong2025deepeyesv2}.

% 过一遍现有agentic vlm工作，带一下process reward工作（末尾用一句带一下现有工作问题）
In the text domain, large language models are trained via reinforcement learning (RL) to dynamically acquire external knowledge through search during reasoning~\cite{jin2025searchr1,song2025r1searcher}. Extending this agentic paradigm to the multimodal setting, vision-language models leverage supervised fine-tuning and RL pipelines to master diverse operations, such as text search, image search, cropping, and zooming, to achieve accurate visual question answering~\cite{wu2025mmsearchr1,chng2025sensenovamars,hong2025deepeyesv2,zheng2025deepeyes,liu2025visualarft,li2025pixelreasoner}. However, existing training paradigms typically reward the agent based solely on final answer correctness or mere tool invocation. This outcome-centric supervision serves as an inaccurate proxy for the actual necessity and effectiveness of each tool call.

% 结合图，说明现有方法缺点（重点，相当于摘要however一句的展开），引出需要怎么解决
As illustrated in Figure~\ref{fig:ntepr_overview}, we find that a useful tool invocation is not merely one that happens to appear within a successful reasoning trajectory. Instead, it must function as a complete tool-evidence step, satisfying a dual requirement: it must proactively target necessary evidence prior to execution, and it must successfully extract the required information from the resulting observation afterward. When this rigorous standard is absent, agents are prone to two distinct failure points. \textbf{(i)} Pre-call misalignment: The model may select a off-target redundant call that fails to seek the essential evidence. \textbf{(ii)} Post-acquisition failure: Even when an appropriate tool returns valuable context, the model may fail to extract the necessary evidence required for subsequent reasoning. Consequently, the critical unresolved challenge is evidence-level credit assignment: providing fine-grained supervision to ensure not only that a tool is called, but that the agent's intent is genuinely necessary and the retrieved information is accurately utilized.

% 讲数据构造和reward
To address this gap, we introduce the NTEP (Necessary Tool-Evidence Path), an explicit annotation scheme that specifies the essential external evidence and corresponding tool calls for each training instance. Specifically, an NTEP decomposes the reasoning trajectory into an ordered sequence of answer-critical steps, where each step explicitly defines an evidence-seeking goal, the appropriate tool to execute, and the essential information that must be extracted from the resulting observation. Building upon this, we propose NTEP-R (NTEP Reward), a sample-specific process reward tailored for evidence-seeking tool invocation. During reinforcement learning, NTEP-R provides a dual-phase evaluation for each tool-use step: it rewards the agent for aligning its pre-call intent with a necessary evidence goal, and for ensuring the information summarized from the post-call observation aligns with the required evidence. Furthermore, a non-repeated-goal regularizer penalizes redundant operations that revisit satisfied NTEP goals. This design elevates tool supervision from a coarse binary of whether tools are used, to a fine-grained evaluation of how effectively each call advances the essential evidence trajectory for a given sample.

% 评估
We evaluate NTEP-R across seven image-grounded benchmarks covering search-oriented reasoning and fine-grained visual perception. Operating under a unified three-tool framework (image cropping, visual search, and text retrieval), NTEP-R significantly improves both accuracy and tool-use efficiency over strong end-to-end and agentic baselines. Ablation studies confirm that pre-call goal alignment and post-call information extraction are highly complementary, while our non-repeated-goal regularizer effectively enforces precise, selective tool utilization. Overall, these findings establish that robust agentic VLMs require fine-grained process supervision over evidence acquisition, rather than crude incentives based solely on tool-call volume.

Our contributions are threefold:
\begin{itemize}
    \item We introduce NTEP (Necessary Tool-Evidence Path), a novel formulation that explicitly defines the essential evidence goal, tool action, and required information of each answer-critical step, addressing the lack of fine-grained intentionality in existing outcome-driven methods.
    \item We propose NTEP-R (NTEP Reward), a fine-grained process-reward framework for tool-using agents. It provides a dual-phase supervision that rewards pre-call goal alignment and post-call information extraction, while introducing a non-repeated-goal regularizer to explicitly penalize redundant tool invocations.
    \item We conduct extensive evaluations on seven image-grounded benchmarks under a unified three-tool framework. Our results demonstrate significant improvements in search-oriented accuracy and tool-use efficiency, with comprehensive ablation studies validating the essential contribution of each reward component.
\end{itemize}

\section{Related Work}

\noindent\textbf{Agentic VLM Tool Use.}
Tool-using agents extend language and vision-language models with external actions for search, visual inspection, browsing, and multi-step evidence gathering.
In the text domain, Search-R1, R1-Searcher, R1-Searcher++, StepSearch, and WebThinker show that LLMs can learn to acquire external knowledge through search during reasoning~\cite{jin2025searchr1,song2025r1searcher,song2025r1searcherpp,zheng2025stepsearch,li2025webthinker}.
In the multimodal domain, MMSearch-R1 and SenseNova-MARS train agents with image search, text search, and visual operations, while MM-DeepResearch and WebWatcher scale multimodal search and deep-research workflows~\cite{wu2025mmsearchr1,chng2025sensenovamars,yao2026mm,geng2025webwatcher}.
Another line focuses on visual operations and interactive image reasoning, including DeepEyes, DeepEyesV2, Vision-R1, Visual-ARFT, PixelReasoner, Thyme-RL, Chain-of-Focus, V-Thinker, VISTA-R1, and IMAgent~\cite{zheng2025deepeyes,hong2025deepeyesv2,huang2025visionr1,liu2025visualarft,li2025pixelreasoner,zhang2025thyme,zhang2025chain,qiao2025v,lu2025vistarl,dong2025imagent}.
These pipelines demonstrate that VLMs can learn to interact with tools, but they mainly study whether the agent can complete the task or learn a useful tool behavior, rather than explicitly specifying the evidence role of each call.

\noindent\textbf{Reinforcement Learning for Agent-Tool Interaction.}
Reinforcement learning trains agents to invoke tools with outcome, cost, or process rewards.
Search-agent methods use RL to encourage query issuing, dynamic knowledge acquisition, or step-wise search behavior~\cite{jin2025searchr1,song2025r1searcher,song2025r1searcherpp,zheng2025stepsearch}, and multimodal systems extend this idea to image-grounded search and visual operations~\cite{wu2025mmsearchr1,chng2025sensenovamars,lu2025vistarl}.
Reward methods include ToolRL for tool-learning rewards, RLTR for good processes without correct outcomes, Atom-Searcher for atomic reasoning rewards, and TA-MDP for LVLM reward decomposition~\cite{qian2025toolrl,li2025rltr,deng2025atomsearcher,adams2026tamdp}.

\begin{figure*}[t]
\centering
\setlength{\fboxsep}{4pt}
% \fbox{
% \begin{minipage}[c][1.85in][c]{0.96\textwidth}
% \centering
% \textbf{Placeholder: Two-Stage Training with \methodname{}}\\[2mm]
% \small
% \begin{tabular}{c@{\hspace{0.45em}}c@{\hspace{0.45em}}c@{\hspace{0.45em}}c@{\hspace{0.45em}}c@{\hspace{0.45em}}c@{\hspace{0.45em}}c}
% \fbox{\begin{minipage}[c][0.72in][c]{0.17\textwidth}
% \centering
% \textbf{Offline Construction}\\[-0.5mm]
% \scriptsize
% target-policy warm-up rollouts\\
% teacher extracts or completes
% \end{minipage}}
% &
% $\Longrightarrow$
% &
% \fbox{\begin{minipage}[c][0.72in][c]{0.15\textwidth}
% \centering
% \textbf{Frozen NTEP}\\[-0.5mm]
% \scriptsize
% answer-critical transitions\\
% $g_j \xrightarrow{\,t_j\,} e_j$
% \end{minipage}}
% &
% $\Longrightarrow$
% &
% \fbox{\begin{minipage}[c][0.72in][c]{0.20\textwidth}
% \centering
% \textbf{GRPO Rollout Scoring}\\[-0.5mm]
% \scriptsize
% $R_{\mathrm{ans}} + R_{\mathrm{fmt}}$\\
% \textsc{Align} before call\\
% \textsc{Acquire} after return
% \end{minipage}}
% &
% $\Longrightarrow$
% &
% \fbox{\begin{minipage}[c][0.72in][c]{0.15\textwidth}
% \centering
% \textbf{Policy Update}\\[-0.5mm]
% \scriptsize
% composite reward\\
% update $\pi_\theta$
% \end{minipage}}
% \end{tabular}\\[3mm]
% \scriptsize
% \textit{At inference, the learned policy interacts with the tools directly, without the teacher, semantic judge, or NTEP annotations.}
% \end{minipage}}
\includegraphics[width=\textwidth]{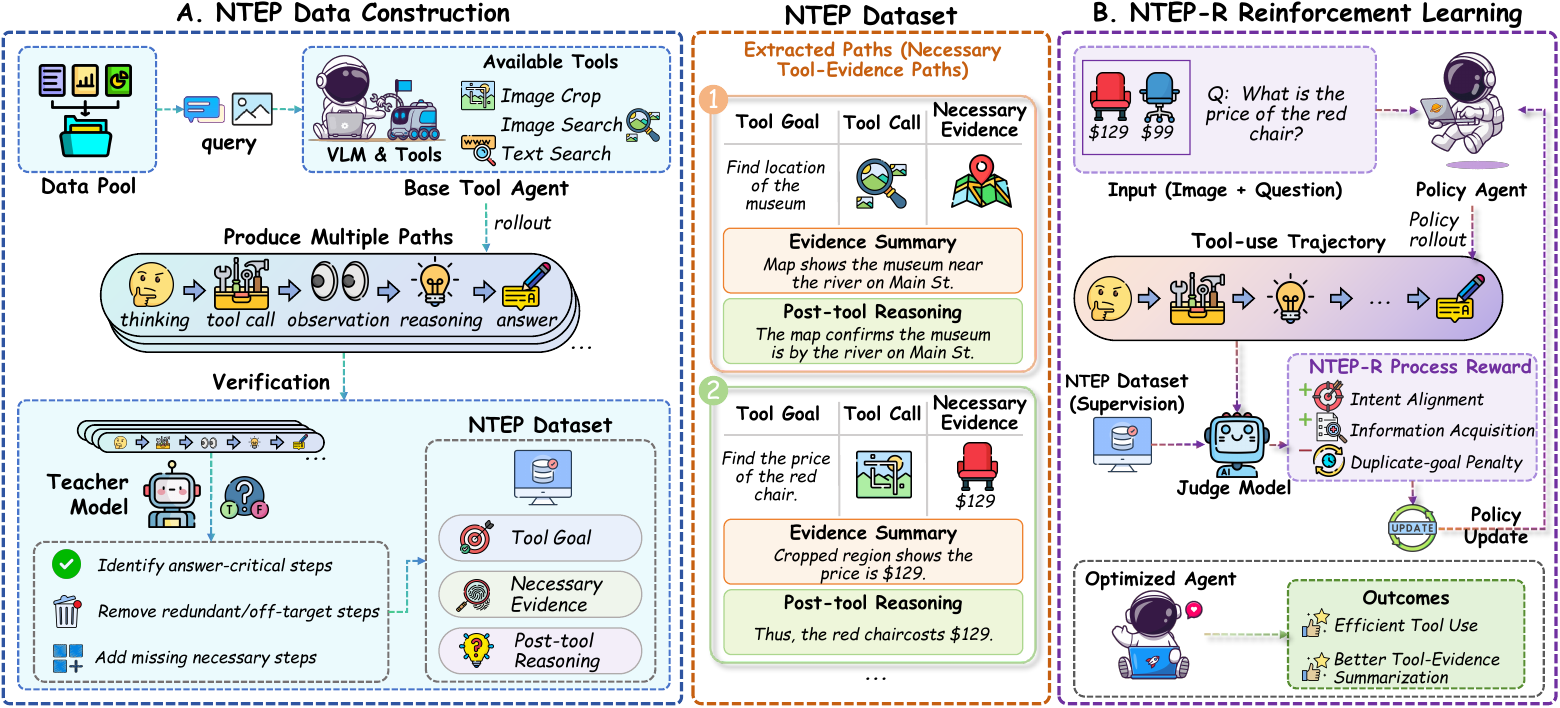}
\caption{\textbf{Two-stage overview of \methodname{}.} Before the main GRPO stage, a teacher distills target-policy warm-up rollouts into frozen, sample-specific NTEPs. During GRPO, current-policy rollouts receive verifiable answer and format rewards, while a semantic judge evaluates pre-call goal alignment and post-call information acquisition against the NTEP. \methodname{} aggregates these signals to update the policy.}
\label{fig:method_overview}
\end{figure*}

\section{Method}

\subsection{Problem Formulation}

We consider a general agentic vision-language workflow in which answering a question may require evidence beyond the initial visual input.
Given an input $x=(v,q)$ containing a visual context $v$ and a question $q$, a policy $\pi_\theta$ may interact with a set of external tools $\mathcal{T}$ before producing an answer $y$.
At interaction step $i$, the policy forms a reasoning state or intent $r_i$, issues a call $c_i=(t_i,a_i)$ with tool $t_i\in\mathcal{T}$ and arguments $a_i$, receives an observation $o_i$, and continues reasoning.
This yields a trajectory
\begin{equation}
\tau = (r_1, c_1, o_1, r_2, c_2, o_2, \ldots, r_K, c_K, o_K, y),
\label{eq:trajectory}
\end{equation}
where $K$ is not fixed and $\mathcal{T}$ may include visual operations, retrieval systems, OCR, code interpreters, databases, or other external modules.

\noindent\textbf{Objective.}
The objective is to produce a correct answer while acquiring the necessary evidence through valid and efficient interactions.
The central problem is to identify which intermediate interactions contribute necessary evidence and to assign them process-level credit without assuming a particular tool set or a single canonical tool sequence.
Terminal correctness alone cannot provide this distinction: trajectories with the same answer may contain useful, redundant, or off-target calls, while partially successful trajectories may acquire valid evidence before arriving at an incorrect answer.

\subsection{Overview}

Our key insight is that a valid tool invocation requires process supervision at two complementary stages: (1) verifying whether the agent's pre-call intent targets the necessary evidence, and (2) ensuring its post-call reasoning successfully extracts this evidence from the observation. Rather than enforcing rigid reasoning trajectories, NTEPs define these essential evidence milestones for each training sample, granting the agent flexibility in formulating exact tool arguments and intermediate steps.
As illustrated in Figure~\ref{fig:method_overview}, a teacher first distills target-policy warm-up rollouts into frozen NTEPs; during GRPO, verifiable outcome rewards and NTEP-conditioned semantic judgments jointly score current-policy trajectories.
\methodname{} turns these signals into process-level credit while penalizing redundant calls that revisit the same NTEP goal.

\subsection{Necessary Tool-Evidence Paths}

Concretely, we model each necessary step as a tool-mediated evidence transition,
\begin{equation}
g_j \xrightarrow{\,t_j\,} e_j,
\label{eq:ntep_step}
\end{equation}
where $g_j$ is the evidence goal before the call, $t_j \in \mathcal{T}$ is the tool expected to acquire that evidence, and $e_j$ is the necessary information that should be extracted from the returned observation.
For an input $x$, a Necessary Tool-Evidence Path (NTEP) is the ordered collection of these answer-critical transitions,
\begin{equation}
\mathcal{P}(x)=\bigl((g_j,t_j,e_j)\bigr)_{j=1}^{N}.
\label{eq:ntep}
\end{equation}
By defining essential evidence targets rather than prescribing exact wording or rigid trajectories, NTEPs allow multiple valid queries and reasoning traces to achieve the same goal. Importantly, this abstraction is tool-agnostic. A new workflow can seamlessly incorporate any tool $t_j$ from its own inventory, provided its pre- and post-action states are mapped to our universal reward interface.

We construct NTEPs from the target agent's own warm-up rollouts and use the teacher as a distiller rather than a trajectory demonstrator.
Target rollouts accurately reflect workflow-specific evidence needs that a stronger teacher might skip, though they often contain redundant or failed interactions.
To resolve this, the teacher filters answer-critical transitions from successful rollouts and imputes missing evidence steps into unsuccessful ones, producing standardized $(g_j, t_j, e_j)$ targets in both cases.

\subsection{NTEP Reward}

To ensure that every tool invocation actively advances the reasoning process, NTEP Reward (NTEP-R) translates the predefined evidence paths in two complementary judgments around each tool interaction.
Let $r_i^{+}$ denote the first reasoning state after observation $o_i$ is returned.
For each still-unfinished NTEP step, we define
\begin{equation}
\begin{aligned}
\alpha_{i,j} &= \operatorname{Align}(r_i,c_i;g_j,t_j), \\
\beta_{i,j} &= \operatorname{Acquire}(r_i^{+};e_j),
\end{aligned}
\qquad \alpha_{i,j},\beta_{i,j}\in\{0,1\}.
\label{eq:ntep_judgments}
\end{equation}
Here, \textsc{Align} asks whether the pre-call intent and the tool call target the required evidence goal, while \textsc{Acquire} asks whether the post-call reasoning explicitly captures the necessary information.
They are independent binary semantic judgments from a frozen training-time judge, used only for reward computation.
Scoring them separately localizes credit to evidence-seeking and evidence uptake instead of collapsing both into terminal-answer supervision.

Given an agent's reasoning trajectory $\tau$ and its corresponding reference path $\mathcal{P}(x)$, we sequentially evaluate each tool invocation to track its progress against the pending NTEP. During this process, we record four key metrics:
\begin{align}
H_{\mathrm{goal}}
&= \sum_{j=1}^{N} \mathbb{I}\!\left[\sum_{i=1}^{K} \alpha_{i,j} \ge 1\right], \\
H_{\mathrm{info}}
&= \sum_{j=1}^{N} \mathbb{I}\!\left[\sum_{i=1}^{K} \beta_{i,j} \ge 1\right], \\
M_{\mathrm{goal}}
&= \sum_{i=1}^{K} \mathbb{I}\!\left[\max_j \alpha_{i,j}=0\right], \\
D_{\mathrm{dup}}
&= \sum_{j=1}^{N} \max\!\left(0,\sum_{i=1}^{K}\alpha_{i,j}-1\right),
\end{align}
where $H_{\mathrm{goal}}$ denotes distinct evidence goals targeted by the agent; $H_{\mathrm{info}}$ represents necessary information items extracted; $M_{\mathrm{goal}}$ tracks off-path calls matching no required goal; and $D_{\mathrm{dup}}$ counts redundant calls that retarget a goal already satisfied earlier.
Thus, $H_{\mathrm{goal}}$ and $H_{\mathrm{info}}$ credit evidence-seeking intent and post-call information uptake, while the remaining terms discourage off-path and repetitive actions.

The NTEP-R process reward is
\begin{equation}
\begin{aligned}
R_{\mathrm{NTEP}}(\tau,x)
=
\frac{1}{N}\bigl(
&(1-\lambda_g)H_{\mathrm{info}}
+\lambda_g H_{\mathrm{goal}} \\
&-\lambda_g M_{\mathrm{goal}}
-\lambda_d D_{\mathrm{dup}}
\bigr),
\end{aligned}
\label{eq:ntep_reward}
\end{equation}
where $\lambda_g$ balances information acquisition against goal alignment, and $\lambda_d$ controls the non-repeated-goal regularizer.
Only the first call aligned with a goal receives alignment credit; each additional call aligned with that goal is treated as redundant and explicitly penalized.
Removing this regularizer yields the corresponding component ablation.

\begin{table*}[t!]
\centering
\small
\setlength{\tabcolsep}{2.6pt}
\begin{tabular}{@{}l ccccc cccc c@{}}
\toprule
& \multicolumn{5}{c}{Search-oriented (\%)} & \multicolumn{4}{c}{Visual (\%)} & \multicolumn{1}{c}{Avg. (\%)} \\
\cmidrule(lr){2-6}\cmidrule(lr){7-10}
Method & MMSe & HR-MMSe & InfoS & MAT & S-Avg.$\uparrow$ & V* & HR-4K & HR-8K & V-Avg.$\uparrow$ & Avg.$\uparrow$ \\
\midrule
\rowcolor{GroupRow}
\multicolumn{11}{@{}l}{\textit{(i) End-to-end VLMs (no tools)}} \\
Qwen3-VL-8B~\cite{bai2025qwen3vl} & 11.11 & 2.95 & 21.55 & 62.00 & 24.40 & 87.43 & 82.62 & 74.25 & 81.44 & 48.85 \\
Qwen3-VL-30B~\cite{bai2025qwen3vl} & 15.79 & 3.28 & 27.05 & 66.00 & 28.03 & 86.39 & 79.88 & 74.88 & 80.38 & 50.47 \\
GPT-5~\cite{openai2026gpt5} & 38.01 & 14.43 & 50.30 & 78.67 & 45.35 & 72.77 & 75.75 & 73.50 & 74.01 & 57.63 \\
Claude-Sonnet-4.6~\cite{anthropic2026sonnet46} & 32.75 & 12.13 & 49.80 & 74.67 & 42.34 & 71.73 & 76.12 & 68.75 & 72.20 & 55.14 \\
Grok-4~\cite{xai2025grok4} & 36.26 & 17.38 & 40.00 & 86.00 & 44.91 & 82.20 & 79.25 & 73.88 & 78.44 & 59.28 \\
Gemini-2.5-Pro~\cite{comanici2025gemini25} & 35.67 & 12.13 & 54.55 & 81.33 & 45.92 & 81.15 & 86.62 & 81.50 & 83.09 & 61.85 \\
\midrule
\rowcolor{GroupRow}
\multicolumn{11}{@{}l}{\textit{(ii) RL agent pipelines (three-tool evaluation)}} \\
Vision-R1-7B~\cite{huang2025visionr1} & 17.5 & 4.6 & 23.4 & 57.3 & 25.70 & 76.4 & 72.4 & 69.5 & 72.77 & 45.87 \\
MM-DeepResearch-8B~\cite{yao2026mm} & 33.3 & 17.4 & 29.8 & 77.3 & 39.46 & 56.5 & 68.1 & 56.9 & 60.51 & 48.48 \\
DeepEyes-7B~\cite{zheng2025deepeyes} & 31.0 & 11.5 & 22.9 & 68.0 & 33.34 & 71.2 & 66.2 & 57.4 & 64.94 & 46.89 \\
DeepEyesV2-7B~\cite{hong2025deepeyesv2} & 24.6 & 10.8 & 29.6 & 68.0 & 33.23 & 66.0 & 64.4 & 56.4 & 62.24 & 45.66 \\
Thyme-RL~\cite{zhang2025thyme} & 15.2 & 3.9 & 25.1 & 61.3 & 26.38 & 70.2 & 66.1 & 61.1 & 65.80 & 43.28 \\
Chain-of-Focus-7B~\cite{zhang2025chain} & 21.1 & 4.9 & 21.9 & 56.0 & 25.97 & 69.1 & 60.4 & 52.6 & 60.70 & 40.85 \\
V-Thinker-7B~\cite{qiao2025v} & 16.4 & 5.2 & 24.4 & 51.3 & 24.35 & 71.2 & 61.2 & 55.2 & 62.57 & 40.73 \\
PixelReasoner-7B~\cite{li2025pixelreasoner} & 18.1 & 5.2 & 23.5 & 58.7 & 26.39 & 57.1 & 57.9 & 51.5 & 55.48 & 38.85 \\
Visual-ARFT-Search~\cite{liu2025visualarft} & 24.6 & 9.2 & 25.6 & 70.0 & 32.35 & 42.9 & 46.5 & 41.2 & 43.56 & 37.15 \\
WebWatcher-7B~\cite{geng2025webwatcher} & 24.0 & 17.0 & 29.8 & 63.3 & 33.54 & 32.5 & 37.0 & 31.4 & 33.61 & 33.57 \\
SenseNova-MARS-8B~\cite{chng2025sensenovamars} & \underline{62.0} & \underline{34.1} & \underline{54.4} & \textbf{82.0} & \underline{58.12} & \textbf{92.7} & \underline{78.2} & \underline{74.8} & \underline{81.90} & \underline{68.31} \\
\midrule
\rowcolor{OursRow}
\textbf{\modelname{} (ours)} & \textbf{68.4} & \textbf{35.7} & \textbf{56.8} & \underline{81.3} & \textbf{60.55} & \underline{90.6} & \textbf{81.2} & \textbf{78.4} & \textbf{83.40} & \textbf{70.34} \\
\bottomrule
\end{tabular}
\caption{\textbf{Main results.} S-Avg., V-Avg., and Avg. macro-average search, visual, and all splits; \textbf{bold}/underline mark best/second-best RL-agent results.}
\label{tab:main_results}
\end{table*}

\subsection{Reinforcement Learning with \methodname{}}

Because each tool observation dynamically updates the context for future decisions, trajectory-level optimization is better suited for tool use than isolated behavior cloning. Since NTEPs define evidence milestones rather than rigid action sequences, standard imitation learning would unnecessarily constrain the agent's exploration space. To address this, we leverage GRPO~\cite{shao2024deepseekmath} to optimize over alternative trajectories, prioritizing those that efficiently meet all evidence requirements. Crucially, we retain GRPO's clipped policy update but modify its group-relative advantage computation to operate at the token level, parameterized by our path-conditioned composite reward.

\noindent\textbf{Interaction Protocol.}
We utilize an XML-style protocol to explicitly expose the state transitions required for reward computation:
\begin{itemize}
    \item \textbf{Pre-call State} (\textsc{Align}): The $\texttt{<thinking>}$ tag captures the intent $r_i$, followed by a JSON action $c_i=(t_i,a_i)$ inside $\texttt{<tool\_call>}$.
    \item \textbf{Post-call State} (\textsc{Acquire}): The environment injects the observation $o_i$ into \texttt{<tool\_response>}, and the agent's subsequent $\texttt{<thinking>}$ step $r_i^{+}$ integrates this information.
    \item \textbf{Termination}: The $\texttt{<answer>}$ block provides the non-empty final prediction $y$ and halts the trajectory.
\end{itemize}

To enforce adherence to this schema, the format reward $R_{\mathrm{fmt}}$ is strictly positive if and only if the structural syntax is valid, all JSON calls contain registered tools with legitimate arguments, and no model-induced execution errors occur.

\noindent\textbf{Trajectory Sampling and Reward Construction.}
For each training input $x$, the current policy samples a group of $G$ trajectories $\{\tau_g\}_{g=1}^{G}$.
Each trajectory receives verifiable answer and format rewards together with the process reward from Equation~\ref{eq:ntep_reward}.
The composite reward is
\begin{equation}
R_g
= R_{\mathrm{ans}}(y_g,y^\ast)
+ R_{\mathrm{fmt}}(\tau_g)
+ \lambda R_{\mathrm{NTEP}}(\tau_g,x),
\label{eq:total_reward}
\end{equation}
where $\lambda$ controls the contribution of semantic process supervision.
The answer reward anchors task success, while the process term distinguishes trajectories by how effectively they acquire the required evidence.

\noindent\textbf{Group-Relative Optimization.}
For the trajectories sampled for $x$, let $\bar{R}_x$ and $s_x$ denote the mean and standard deviation of their composite rewards.
We define the NTEP-conditioned group-relative advantage as
\begin{equation}
\widehat{A}^{\mathrm{NTEP}}_g
= \frac{R_g-\bar{R}_x}{s_x+\epsilon}.
\label{eq:ntep_advantage}
\end{equation}
Through Eq.~\ref{eq:total_reward}, the NTEP process signal enters $R_g$ before group normalization, turning outcome-centered GRPO into a path-conditioned learning signal.
Because trajectories in the group share the same input and frozen NTEP, this advantage compares not only task outcomes but also how effectively each trajectory acquires the required evidence.
Unlike outcome-only group scoring, answer-equivalent trajectories can receive different learning signals according to their goal alignment, information uptake, and redundant actions.
We use $\widehat{A}^{\mathrm{NTEP}}_g$ in the standard clipped GRPO objective without introducing a separate critic.

\noindent\textbf{Inference.}
The teacher, semantic judge, and NTEP annotations are used only during training.
At test time, the policy receives only the image, question, and tool interface and interacts independently, without a scaffold or oracle path.

\begin{figure*}[t]
\centering
\includegraphics[width=0.98\textwidth]{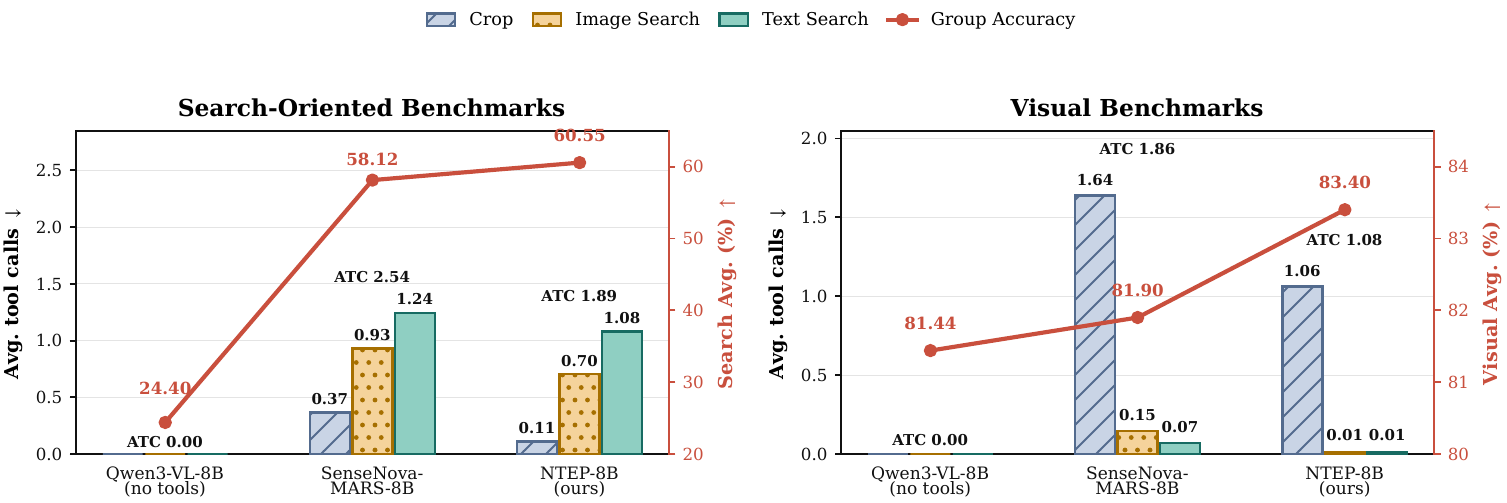}
\caption{Accuracy and tool-call profiles under the main-table protocol. Bars show average calls by tool (left axes), ATC labels give total calls, and lines show accuracy (right axes). Qwen3-VL-8B is end-to-end; RL agents use the common three-tool interface.}
\label{fig:main_tool_use}
\end{figure*}

\section{Experiments}

\subsection{Experimental Setup}

\noindent\textbf{Datasets.}
We evaluate on seven image-grounded benchmarks.
MMSearch~\cite{jiang2025mmsearch}, HR-MMSearch~\cite{chng2025sensenovamars}, InfoSeek~\cite{chen2023infoseek}, and MAT-Search~\cite{liu2025visualarft} evaluate search-oriented reasoning and external evidence acquisition; V* Bench~\cite{wu2024vstar} and HR-Bench 4K/8K~\cite{wang2025hrbench} evaluate fine-grained and high-resolution visual perception.
The main comparison uses matched 500-example subsets for InfoSeek and HR-Bench 4K/8K when comparing \modelname{} with SenseNova-MARS-8B, and full splits for MMSearch, HR-MMSearch, MAT-Search, and V*; end-to-end and external-agent rows use full splits.
Component ablation and budget sweep use all 4,417 examples across the seven evaluation splits.

\noindent\textbf{Baselines.}
We compare two groups in Table~\ref{tab:main_results}: end-to-end VLMs that answer without tools, and RL agent pipelines that perform multi-turn tool interaction.
The former include Qwen3-VL-8B/30B~\cite{bai2025qwen3vl}, GPT-5~\cite{openai2026gpt5}, Grok-4~\cite{xai2025grok4}, Gemini-2.5-Pro~\cite{comanici2025gemini25}, and Claude-Sonnet-4.6~\cite{anthropic2026sonnet46}; the latter include \modelname{}, SenseNova-MARS-8B~\cite{chng2025sensenovamars}, Vision-R1-7B~\cite{huang2025visionr1}, and the external agent checkpoints listed in the table.
End-to-end VLMs use no tools.
RL-agent rows use a common three-tool protocol; external checkpoints are evaluated through a generic adapter exposing the same tools, so these rows measure transfer to a common interface rather than native-pipeline performance.

\noindent\textbf{Metrics.}
We report Accuracy (Acc.) and Average Tool Calls (ATC), split into Crop, Image, Text, and Total calls.
Search Average (S-Avg.), Visual Average (V-Avg.), and Overall Average (Avg.) macro-average the four search, three visual, and all splits from unrounded scores.
Trajectory and failure-mode analyses use 300 stratified judge-labeled examples per model.

\noindent\textbf{Implementation Details.}
\modelname{} and its ablations are initialized from Qwen3-VL-8B-Instruct~\cite{bai2025qwen3vl} and use image crop, image search, and text search with a shared ten-call budget.
Training follows the two-stage procedure in Figure~\ref{fig:method_overview}; during GRPO~\cite{shao2024deepseekmath}, we sample $G=8$ trajectories per input with a 32K context window and a maximum 16K response length.
All same-harness comparisons use matched benchmark splits and serving configurations.
Unless otherwise stated, each reported checkpoint result is from one training run and one evaluation pass under the fixed decoding seed.
Closed-source baselines receive the same prompt, full-resolution image input, and requested decoding settings; unsupported API parameters are omitted and recorded.

\subsection{Main Results}

\noindent\textbf{Best same-harness RL agent.}
Table~\ref{tab:main_results} compares all methods on seven image-grounded benchmarks.
Within the unified three-tool harness, \modelname{} obtains the best RL-agent average of 70.34, exceeding the reproduced SenseNova-MARS-8B checkpoint by 2.03 points.
The gain is not concentrated in one benchmark group: Search Avg. improves from 58.12 to 60.55, and Visual Avg. improves from 81.90 to 83.40.
Tool-using agents are especially important on search-oriented tasks: the strongest end-to-end baseline, Gemini-2.5-Pro, reaches 45.92 Search Avg., while \modelname{} reaches 60.55 under the same protocol.
This gap indicates that search-heavy tasks still require active evidence acquisition beyond parametric knowledge and full-resolution image input.

\noindent\textbf{Tool-Use Efficiency.}
Figure~\ref{fig:main_tool_use} makes the corresponding operating points explicit. Compared to SenseNova-MARS-8B, \modelname{} reduces average tool calls from $2.54$ to $1.89$ on search-oriented benchmarks and from $1.86$ to $1.08$ on visual benchmarks, while improving accuracy in both groups.
The gain therefore comes from more selective tool use rather than more frequent invocation.

\subsection{Tool-call Analysis}
\begin{figure}[t]
\centering
\includegraphics[width=\columnwidth]{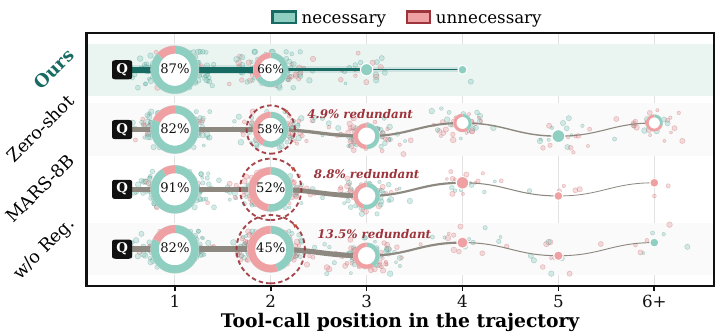}
\caption{Tool-call trajectories. Ring area encodes frequency; mint/red splits mark necessary-and-used versus unnecessary calls.}
\label{fig:journey}
\end{figure}

\begin{figure}[t]
\centering
\includegraphics[width=\columnwidth]{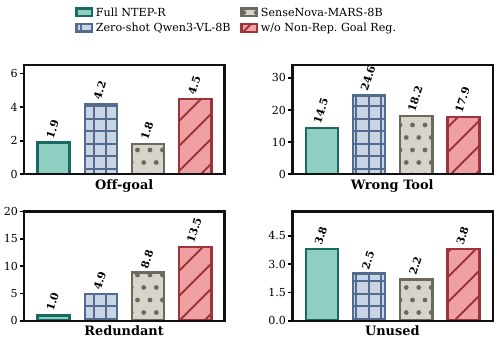}
\caption{Failure modes over calls from 300 judged trajectories/model (\%; lower is better).}
\label{fig:failure_modes}
\end{figure}

\noindent\textbf{Tool-call trajectory.}
The left panel of Figure~\ref{fig:journey} shows where tool calls occur.
\modelname{} concentrates evidence acquisition in the first two positions and never reaches a fifth call in this sample.
The baselines continue into later positions, where useful-call rates decay and repeated evidence acquisition becomes more visible.
The same audit gives \modelname{} a $78.8\%$ necessary-and-used call rate, compared with $69.0\%$ for SenseNova-MARS-8B; the stricter Case-5 rate, requiring every call to be necessary and the final answer to be correct, rises from $39.0\%$ to $58.7\%$.

\noindent\textbf{Failure modes.}
To localize residual waste, we group non-necessary calls in the judged sample into four disjoint failure modes: off-goal targeting, wrong-tool selection, repeated evidence acquisition, and evidence returned but unused.
For deterministic labeling, categories follow the priority redundant $>$ off-goal $>$ wrong tool $>$ unused.
Figure~\ref{fig:failure_modes} shows that the regularizer acts as intended: redundancy falls from $13.5\%$ without regularization to $1.0\%$ in \modelname{}.
Wrong-tool selection remains the largest residual failure for every model, suggesting tool-choice supervision as a complementary direction beyond evidence-path supervision.

\begin{figure}[t!]
\centering
\includegraphics[width=\columnwidth]{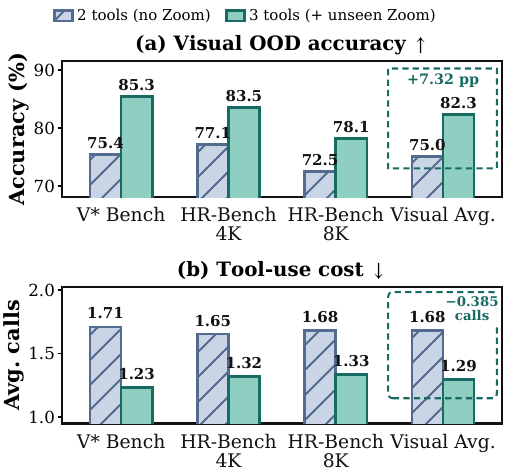}
\caption{Cross-Tool OOD transfer. Adding unseen zoom to search-only interface improves accuracy and reduces ATC.}
\label{fig:ood_visual_toolset}
\end{figure}

\subsection{Cross-Tool OOD Transfer}
The NTEP-R objective trains the policy around abstract evidence states rather than overfitting to the training tool distribution. To assess this out-of-distribution (OOD) transfer, we use a policy trained exclusively on search tasks, completely omitting the V* and HR-Bench datasets, as well as the visual zoom interface. We then evaluate this model on the held-out visual benchmarks under two settings: a restricted search-only interface, and an extended interface incorporating the unseen zoom tool. As shown in Figure~\ref{fig:ood_visual_toolset}, without any further RL updates, activating this OOD tool raises the Visual Avg. accuracy from $75.01\%$ to $82.32\%$ while reducing average calls from $1.679$ to $1.294$ per query. This simultaneous gain in accuracy and tool-use efficiency confirms that the evidence-seeking strategy learned during search training successfully transfers to visual operations: the model efficiently invokes the novel zoom tool to acquire missing details and halts once the goal is met. Appendix~\ref{app:theory} provides a formal theoretical account of this transfer behavior.

\subsection{Ablation Studies}

\begin{figure}[t]
\centering
\includegraphics[width=\columnwidth]{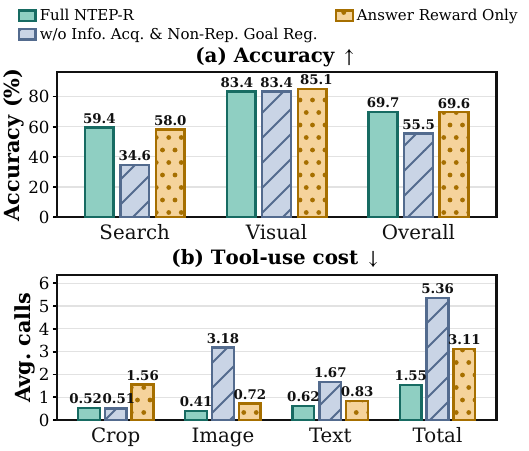}
\caption{Reward-component ablation. \textbf{Top:} split-macro accuracy. \textbf{Bottom:} average calls by tool and total.}
\label{fig:ablation_components}
\end{figure}

Figure~\ref{fig:ablation_components} evaluates the reward design from both accuracy and call-efficiency perspectives.
All variants keep the shared format reward; the labels only describe which NTEP process terms are removed.
The w/o Information Acquisition and Non-Repeated-Goal Regularizer arm keeps only pre-call goal alignment and is shown at its final collapsed checkpoint, while Answer Reward Only is a smaller-pool legacy diagnostic rather than a strict single-variable run.

\noindent\textbf{Accuracy.}
Goal alignment by itself is insufficient: without information acquisition and duplicate-goal regularization, Search Avg. collapses to 34.58 even though the policy still learns to issue many calls.
Full \methodname{} restores Search Avg. to 59.44 and Overall Avg. to 69.69, while leaving Visual Avg. essentially unchanged at 83.36 versus 83.37.
Answer-reward-only RL reaches a similar Overall Avg. of 69.63, but this number hides substantially different tool behavior.

\noindent\textbf{Calls.}
The call counts expose the difference that accuracy alone misses.
The goal-only variant averages 5.36 calls per sample, and Answer Reward Only still uses 3.11 calls.
Full \methodname{} reduces this to 1.55 calls, cutting total calls by roughly $71\%$ relative to the goal-only variant and by roughly $50\%$ relative to Answer Reward Only.

\subsection{Judge Reliability}

Automated judgments are used only for open-ended search-answer scoring and trajectory-level analysis labels; multiple-choice visual benchmarks use exact matching.
We therefore audit a random sample covering all judge-involved settings with blinded human adjudication; Appendix~\ref{app:human_audit} gives the full protocol and disagreement breakdown.
Table~\ref{tab:judge_reliability} summarizes the protocol and agreement.
The high agreement supports using the frozen judge as a reliable measurement and reward signal, while keeping visual benchmark scores independent through exact matching.

\begin{table}[t]
\centering
\small
\setlength{\tabcolsep}{3.8pt}
\renewcommand{\arraystretch}{1.04}
\begin{tabular}{@{}p{0.24\columnwidth}p{0.66\columnwidth}@{}}
\toprule
Item & Value \\
\midrule
Scope & 100 blinded cases covering search-answer scoring and trajectory labels \\
Inputs & Image, question, model answer, tool transcript, and returned evidence \\
Labels & Correctness, tool necessity, and failure mode for non-necessary calls \\
Use & Reliability check only; no model selection or reward tuning \\
Agreement & 97.0\%; Cohen's $\kappa=0.94$ \\
\bottomrule
\end{tabular}
\caption{Human audit of judge reliability.}
\label{tab:judge_reliability}
\end{table}

\section{Conclusion}

We presented \methodname{}, a reward framework that trains agentic VLMs to follow necessary tool-evidence paths rather than optimize only final-answer correctness.
By supervising both pre-call goal alignment and post-call information acquisition, and by regularizing repeated evidence-seeking goals, \methodname{} improves the selectivity of tool use in a unified crop, image-search, and text-search environment.
Across seven benchmarks, \modelname{} improves over the strongest same-harness RL-agent baseline while reducing average tool calls, and our analyses show that the gains come from earlier, more necessary, and less redundant evidence acquisition.
The visual OOD transfer results suggest that \methodname{} learns reusable evidence-seeking discipline rather than memorizing a fixed tool inventory.
Overall, explicit evidence-path supervision is a practical direction for VLM agents that are more accurate, efficient, and auditable.

\clearpage
\bibliography{aaai2027}
\clearpage
\appendix

\section{Additional Methodological Details}
\label{app:method_details}

\subsection{NTEP Data Fields}

Each NTEP annotation is stored as three aligned lists: \texttt{target\_list}, \texttt{tool\_list}, and \texttt{info\_list}.
The $j$-th entries correspond to the evidence goal $g_j$, expected tool $t_j$, and necessary information $e_j$ in Equation~\ref{eq:ntep_step}.
This representation is intentionally looser than a full trajectory transcript: it does not prescribe exact query strings, bounding boxes, or intermediate wording, but records the evidence milestone that a valid trajectory must satisfy.

The construction step uses target-policy warm-up rollouts as the source material.
For successful rollouts, the teacher removes redundant and non-answer-critical interactions and keeps only the evidence-bearing steps.
For unsuccessful rollouts, the teacher identifies the missing evidence transitions needed to complete the question and writes them into the same $(g_j,t_j,e_j)$ schema.
The resulting paths are frozen before GRPO; during RL, the current policy is scored against these paths but does not see them in its prompt.

\subsection{Training Data Composition}
\label{app:data_composition}

Table~\ref{tab:app_data_composition} reports the per-source composition of the two training pools used in this work.
Both pools draw on the same five public sources: FVQA, DeepEyes-4K (split into multiple-choice and open-ended subsets), Visual-Probe, VDR, and the text-only Search-R1 corpus, whose paths contain only \toolcall{text\_search\_tool} steps and are included to preserve strong text-retrieval reasoning.
The historical 7,774-example pool spans all three tools and underlies the reward-composition diagnostics in Appendix~\ref{app:additional_experiments}.
The search-only 4,855-example pool used to train the selected \modelname{} checkpoint keeps exactly those examples whose necessary paths contain no crop/zoom step; its NTEP toolsets decompose into 2,309 text-search-only, 1,058 image-search-only, 1,483 dual-search, and 5 tool-free paths.

The extraction and completion branches contribute 4,819 (62.0\%) and 2,955 (38.0\%) paths, respectively.
The branch mix tracks source difficulty for the warm-up policy: the easier corpora are extraction-dominated, whereas VDR---the hardest visual source---derives 82.0\% of its paths from the completion branch.
Without the completion branch, VDR would contribute almost no NTEP supervision, weakening exactly the multi-step retrieval patterns it is included to teach; conversely, the text-only Search-R1 paths keep the policy proficient at extracting facts from \toolcall{text\_search\_tool} returns.

\begin{table}[t]
\centering
\small
\setlength{\tabcolsep}{4pt}
\renewcommand{\arraystretch}{1.06}
\resizebox{\columnwidth}{!}{%
\begin{tabular}{@{}lrrrr@{}}
\toprule
& \multicolumn{3}{c}{Historical pool} & Search-only pool \\
\cmidrule(lr){2-4}
Source & Extracted & Completed & Total & Total \\
\midrule
FVQA & 1,643 & 249 & 1,892 & 1,850 \\
DeepEyes-4K (MCQ) & 424 & 164 & 588 & 27 \\
DeepEyes-4K (open-ended) & 358 & 58 & 416 & 6 \\
Visual-Probe & 638 & 271 & 909 & 1 \\
VDR & 356 & 1,618 & 1,974 & 976 \\
Search-R1 (text-only) & 1,400 & 595 & 1,995 & 1,995 \\
\midrule
\textbf{Total} & \textbf{4,819} & \textbf{2,955} & \textbf{7,774} & \textbf{4,855} \\
\bottomrule
\end{tabular}}
\caption{Per-source composition of the two training pools. Extracted = paths distilled from correct warm-up rollouts; Completed = paths reconstructed from failed rollouts. The search-only pool retains the examples whose necessary paths involve no crop/zoom step.}
\label{tab:app_data_composition}
\end{table}

\subsection{Tool Interface}

The main experiments use a unified three-tool interface.
The registered tool names are \toolcall{image\_zoom\_in\_tool} for region crop/zoom, \toolcall{image\_search\_tool} for reverse image search, and \toolcall{text\_search\_tool} for textual web/search retrieval.
At inference time, the model receives only the image, the question, and the tool schemas.
It does not receive the NTEP annotation, teacher explanations, or judge feedback.

Concretely, \toolcall{image\_zoom\_in\_tool} crops the referenced region using bounding-box coordinates normalized to $[0,1000]$ and re-encodes the crop with the model's native smart-resize preprocessing before returning it as a new visual observation.
\toolcall{image\_search\_tool} performs reverse image search over the input image and returns the top retrieved entries as titled thumbnails, and \toolcall{text\_search\_tool} sends a textual query to a shared retrieval gateway that returns the top-3 summarized results.
All rollouts operate under a single global budget of at most 10 interaction turns with no per-tool caps, so the composition of zoom and search calls is left entirely to the policy; individual tool responses are truncated to an 8K-token cap.

\subsection{Reward Accounting}

The implementation follows the reward decomposition in Section~3 with the concrete accounting reported as part of the experimental setup in Table~\ref{tab:app_experimental_setup}.
The process score is normalized by the NTEP length $N$ and then multiplied by the process-reward scale used in training.
Only the first semantic hit for a goal receives goal credit; later calls that re-target a still-pending, already-scored goal are counted as duplicates, and calls matching no pending goal are counted as misses.

All semantic judgments---pre-call goal alignment, post-call information acquisition, and open-ended answer scoring---are produced by a frozen Qwen3-VL-Plus judge served behind an OpenAI-compatible endpoint and queried with temperature $0$, an 8K-token output cap, and a structured \texttt{<judge>Yes/No</judge>} verdict format that is parsed deterministically.
The judge is used exclusively inside reward computation; the policy never observes its outputs.

\subsection{Training Hyperparameters}
\label{app:hyperparams}

Table~\ref{tab:app_hyperparams} lists the optimization and serving settings that complement the protocol summary in Table~\ref{tab:app_experimental_setup}.

\begin{table}[t]
\centering
\small
\setlength{\tabcolsep}{3.5pt}
\renewcommand{\arraystretch}{1.08}
\begin{tabular}{@{}p{0.3\linewidth}p{0.64\linewidth}@{}}
\toprule
Component & Setting \\
\midrule
Optimizer & AdamW; lr $1\times10^{-5}$, constant schedule (no warm-up); weight decay $0.01$; gradient clip $1.0$ \\
GRPO sampling & group size $G=8$; batch 128 prompts ($1{,}024$ rollouts per step); mini-batch 128 (single on-policy pass); dynamic micro-batching under a 16K-token cap \\
Policy loss & asymmetric clipping $(0.20,\,0.28)$; token-mean aggregation; entropy coefficient $0$ \\
KL regularization & none (no KL term in reward or loss) \\
Sequence budget & 16K prompt; 16K response; 32K model context; 8K per tool response \\
Rollout sampling & temperature $1.0$; top-$p$ $1.0$; presence penalty $1.5$; fixed seed \\
Rollout serving & asynchronous sglang engine; frozen vision tower \\
Judge serving & Qwen3-VL-Plus; temperature $0$; 8K-token cap; 300\,s timeout; bounded concurrency \\
Path teacher & Gemini-3.1-Pro for both the extraction and completion branches \\
Schedule & 20 epochs; checkpointing and validation every 10 steps \\
Software & Python 3.12; PyTorch 2.9.1; Transformers 5.3.0; sglang 0.5.10; Ray 2.53.0 \\
\bottomrule
\end{tabular}
\caption{Complete training configuration for \modelname{}.}
\label{tab:app_hyperparams}
\end{table}

\subsection{Theoretical Analysis: Call Efficiency and Tool-Interface Transfer}
\label{app:theory}

This subsection formalizes the two mechanisms behind the observations of Sections~4.3--4.4: \methodname{} makes every wasted call individually unprofitable while keeping genuinely needed retries profitable, and it attaches credit to evidence states rather than tool identities, which is what enables transfer to an unseen tool.
We abstract the interaction as an \emph{evidence process}.
For an input $x$ with NTEP $\mathcal{P}(x)$ of length $N$, goal $j$ is \emph{completed} once some call has acquired $e_j$ and \emph{pending} otherwise; $U\subseteq\{1,\dots,N\}$ denotes the pending set.
Verdicts follow the training accounting of Appendix~A.4: calls are judged against the pending steps and each call is attributed to at most one pending step; information credit is granted only through an aligned call's own return, so $\beta_{i,j}\le\alpha_{i,j}$ and an information hit is a conjunction of alignment and acquisition; goal credit is granted on the first alignment of a goal only; a repeated alignment of a pending goal counts in $D_{\mathrm{dup}}$; and a call matching no pending goal counts in $M_{\mathrm{goal}}$.
We use four assumptions.
\textbf{(A1) Measurement locality:} each verdict is a function of the judged call's own reasoning state, invocation, and observation, given the pending set.
\textbf{(A2) Stochastic tool success:} a call aligned with pending goal $g_j$ via tool $t$ acquires $e_j$ with probability $p_t(j)\in(0,1]$. Assuming an optimal policy modulates its arguments (e.g., query reformulations) upon failure, we model these attempts as independent across calls given the goal.
\textbf{(A3) Evidence-conditioned policy:} the policy under analysis selects actions as a function $\pi(a\mid x,U)$ of the input and pending set.
This is the abstraction the process reward is designed to induce---every term of Equation~\ref{eq:ntep_reward} conditions on the goal and evidence semantics $(g_j,e_j)$ of the necessary path, with the expected tool entering only through the \textsc{Align} check, rather than on trajectory templates or the evaluation-time tool inventory---and we analyze policies of this form.
\textbf{(A4) Evidence-monotone answering:} the probability of a correct final answer is nondecreasing in the set of acquired evidence and is unaffected by calls issued after all goals are completed.

\paragraph{Proposition 1 (the reward prices waste, not retries).}
\emph{(i) For any two trajectories $\tau$ and $\tau^{*}$ with the same hit profile $(H_{\mathrm{goal}},H_{\mathrm{info}})$ of which $\tau^{*}$ is waste-free ($M_{\mathrm{goal}}=D_{\mathrm{dup}}=0$),}
\begin{equation}
R_{\mathrm{NTEP}}(\tau^{*})-R_{\mathrm{NTEP}}(\tau)
=\frac{\lambda_g M_{\mathrm{goal}}(\tau)+\lambda_d D_{\mathrm{dup}}(\tau)}{N}:
\label{eq:waste_gap}
\end{equation}
\emph{every off-path call and every redundant alignment carries an individual, additive price.
(ii) The maximal attainable $R_{\mathrm{NTEP}}$ is attained exactly by trajectories that align each goal once and acquire its information on that call: $M_{\mathrm{goal}}=D_{\mathrm{dup}}=0$ and $K=N$ calls.
(iii) Appending, as the final call before answering, a retry of a pending goal $g_j$ whose earlier aligned attempt failed to acquire $e_j$ changes $R_{\mathrm{NTEP}}$ in expectation by $\bigl(p\,(1-\lambda_g)-\lambda_d\bigr)/N$, which is positive whenever $p>\lambda_d/(1-\lambda_g)$ ($=1/7$ at the published constants): the duplicate penalty taxes pure repetition yet leaves genuinely needed retries profitable.}

\noindent\textit{Proof.}
(i) is immediate from Equation~\ref{eq:ntep_reward}: the two trajectories share the credit terms and differ only in the penalty terms.
(ii) Since information credit requires alignment ($\beta\le\alpha$), $H_{\mathrm{info}}\le H_{\mathrm{goal}}\le N$, so $R_{\mathrm{NTEP}}\le\bigl((1-\lambda_g)N+\lambda_g N\bigr)/N$ with equality iff $H_{\mathrm{goal}}=H_{\mathrm{info}}=N$ and $M_{\mathrm{goal}}=D_{\mathrm{dup}}=0$, which forces exactly one aligned, acquiring call per goal.
(iii) The retry adds $(1-\lambda_g)/N$ with probability $p$ (the first acquisition of $e_j$) and costs $\lambda_d/N$ with certainty. \hfill$\square$

\noindent
Equation~\ref{eq:waste_gap} states that the reward prices every wasted call individually, and part (iii) shows the price is calibrated: waste is always unprofitable, while a retry with a realistic success rate remains worthwhile.
Under the group-relative advantage of Equation~\ref{eq:ntep_advantage}, rollouts in a GRPO group that share answer, format, and hit profile differ in composite reward only through the waste term, so the policy gradient moves probability mass from wasteful to waste-free realizations at a rate proportional to their waste gap.
This is the mechanism behind the redundancy collapse from $13.5\%$ to $1.0\%$ (Figure~\ref{fig:failure_modes}) and the uniform call reductions in Figure~\ref{fig:journey}.

\paragraph{Proposition 2 (optimal stopping: call demand is set by the evidence path).}
\emph{Once $U=\varnothing$, every further call earns no credit and strictly lowers the trajectory reward by at least $\min(\lambda_g,\lambda_d)/N$, while by (A4) it cannot raise the answer reward; the reward-optimal policy therefore answers immediately after its last goal completes.
Consequently its behavior is invariant to any budget increase beyond its stopping time $T$, and under (A2) the probability that $T$ exceeds $b$ calls decays geometrically in $b$.}

\noindent\textit{Proof.}
With all goals completed, no call can add goal or information credit (information credit requires alignment with a pending goal), so every further call is a pure penalty---booked as off-path or as a duplicate depending on the accounting---and answering dominates.
Because off-path calls are pure penalties, the reward-optimal policy issues only aligned calls, so $T$ is bounded by a sum of independent geometric variables governed by the selected tools' per-goal success probabilities, whose tail decays geometrically. \hfill$\square$

\noindent
The observed operating point places evaluation squarely in this invariance regime: in an inference-budget sweep over $B\in\{5,10,15,20\}$ on the full 4,417-example suite, Overall Avg.\ moves only within $69.45$--$69.69$ and realized tool use stays within $1.545$--$1.549$ calls per example (Table~\ref{tab:app_budget_sweep}).
The demand for calls is set by the evidence path, not by the budget.

\paragraph{Proposition 3 (evidence-conditioning enables tool-interface transfer).}
\emph{Let $\mathcal{T}\subset\mathcal{T}'$ be tool inventories.
(i) The optimal evidence-conditioned value $V^{*}$ (the supremum of expected composite reward over policies of form (A3)) is monotone in the inventory: $V^{*}_{\mathcal{T}'}(x,U)\ge V^{*}_{\mathcal{T}}(x,U)$ for every state.
The objective's credit is defined by the evidence path rather than by the inventory exposed at evaluation time, so an evidence-conditioned policy has nothing inventory-specific to unlearn when the toolset grows.
(ii) Suppose the policy extends its evidence-conditioned rule greedily to $\mathcal{T}'$: it routes a pending goal $j$ to the highest-success available tool, including the new tool $t^{*}\in\mathcal{T}'\setminus\mathcal{T}$ on the goal class $\mathcal{G}^{*}$ where $p_{t^{*}}(j)>p_{\max}(j)=\max_{t\in\mathcal{T}}p_t(j)$ (e.g., fine-grained visual evidence).
Then for $j\in\mathcal{G}^{*}$ the probability of completing $j$ within $b$ remaining turns rises from $1-(1-p_{\max}(j))^{b}$ to $1-(1-p_{t^{*}}(j))^{b}$, and the expected calls spent on $j$ fall from $\mathbb{E}[\min(\mathrm{Geom}(p_{\max}(j)),b)]$ to $\mathbb{E}[\min(\mathrm{Geom}(p_{t^{*}}(j)),b)]$; with Proposition~2 and (A4), expected accuracy weakly increases while expected calls decrease on instances containing $\mathcal{G}^{*}$ goals.}

\noindent\textit{Proof.}
(i) is feasible-policy inclusion: every policy available under $\mathcal{T}$ remains available under $\mathcal{T}'$, so the supremum cannot decrease.
(ii) The within-budget success probability $1-(1-p)^{b}$ is strictly increasing and the truncated mean $\mathbb{E}[\min(\mathrm{Geom}(p),b)]=(1-(1-p)^{b})/p$ strictly decreasing in $p$ for $b\ge2$; the stopping argument of Proposition~2 applies unchanged to the extended rule---it answers once $U=\varnothing$---so the saved calls are not respent, and (A4) converts additional completed goals into weakly higher answer accuracy. \hfill$\square$

\noindent
The antecedent of (ii)---extending the learned rule to a tool never seen in training, from its interface description alone---relies on the inherent semantic priors of the base large language model. While the RL objective cannot explicitly teach the success probabilities of unseen tools, evidence-conditioned training is precisely what unlocks this capability: by decoupling abstract evidence goals from rigid tool identities, it prevents the policy from overfitting to the training inventory and allows the base model's zero-shot reasoning to naturally map pending visual goals to the novel interface. This property is directly testable.
The theory therefore makes a falsifiable prediction with a distinctive signature: if the rule transfers, accuracy and efficiency must improve \emph{together}, not trade off.
The visual OOD transfer experiment matches this signature on every split (Figure~\ref{fig:ood_visual_toolset}): adding the unseen zoom tool raises Visual Avg.\ from $75.01$ to $82.32$ while cutting average calls from $1.679$ to $1.294$, with per-benchmark values in Appendix~\ref{app:toolset_values}.
An outcome-only objective makes no such prediction: its credit attaches to whole trajectories executed over the training-time inventory, so joint accuracy-and-efficiency gains under an unseen tool are not implied by it.

\section{Case Study}
\label{app:case_study}

\begin{figure*}[t]
\centering
\begingroup
\setlength{\fboxsep}{4.5pt}
\setlength{\fboxrule}{1.0pt}
% Rounded display faces give the diagnostic a light illustrated style;
% observable transcript fields remain in a legible monospace face.
\newcommand{\casecartoon}{\fontfamily{ugq}\fontseries{b}\selectfont}
\newcommand{\caseround}{\fontfamily{qag}\selectfont}
\fcolorbox{black!78}{yellow!13}{%
\begin{minipage}{0.965\textwidth}
\begin{minipage}[c]{0.285\linewidth}
\centering
{\setlength{\fboxsep}{1.6pt}%
\fcolorbox{black!78}{white}{%
\includegraphics[width=0.955\linewidth]{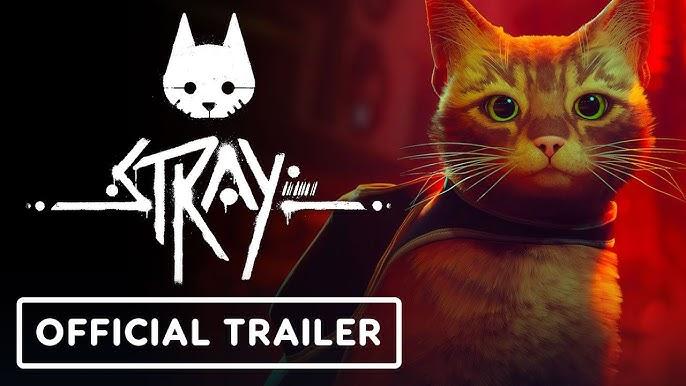}}}\\[-1pt]
{\casecartoon\scriptsize\textcolor{black!70}{REAL MMSEARCH INPUT \(\cdot\) SAMPLE 64}}
\end{minipage}
\hfill
\begin{minipage}[c]{0.675\linewidth}
{\casecartoon\scriptsize\textcolor{blue!65!black}{ORIGINAL QUERY (VERBATIM)}}\\[1pt]
{\caseround\large\textbf{Q:} When is this game come on ninendo switch?}
\par\vspace{5pt}
{\setlength{\fboxsep}{3pt}%
\fcolorbox{DGreen}{white}{%
\begin{minipage}{0.94\linewidth}
{\caseround\small\textcolor{DGreen}{\textbf{Necessary evidence:} \emph{Stray}'s Nintendo Switch release date is November 19, 2024.}}\\[-1pt]
{\casecartoon\scriptsize Ground truth: 2024-11-19}
\end{minipage}}}
\par\vspace{4pt}
{\casecartoon\scriptsize\textcolor{black!65}{SAME SAMPLE \(\cdot\) SAME THREE-TOOL INTERFACE \(\cdot\) SAVED ROLLOUTS}}
\end{minipage}
\end{minipage}}

\vspace{4pt}
{\casecartoon\small\textcolor{black!58}{\(\Downarrow\)\quad REAL, ABRIDGED TRAJECTORY FIELDS\quad \(\Downarrow\)}}
\vspace{4pt}

\fcolorbox{DRed}{DefectRow}{%
\begin{minipage}[t][2.55in][t]{0.455\textwidth}
\begin{minipage}[c]{0.69\linewidth}
{\casecartoon\large\textcolor{DRed}{PATH A: Goal Alignment Only}}\\[-1pt]
{\casecartoon\scriptsize\textcolor{black!62}{CORRECT OUTCOME, REPEATED SATISFIED GOAL}}
\end{minipage}\hfill
\fcolorbox{DRed}{white}{%
\begin{minipage}[c]{0.22\linewidth}
\centering
{\casecartoon\large\textcolor{DRed}{9 CALLS}}\\[-2pt]
{\casecartoon\scriptsize answer judge = 1.0}
\end{minipage}}

\vspace{3pt}\hrule\vspace{4pt}
{\ttfamily\fontsize{6.9}{8.0}\selectfont
\textbf{[Call 1]} text\_search\_tool(\par
\hspace*{1em}query=``Stray game Nintendo Switch release date'')\par
\textbf{[Return 1]} Summary: \ldots{} Nintendo Switch release date for \emph{Stray} is \textbf{November 19, 2024}.\par
\vspace{2pt}
{\casecartoon\scriptsize\textcolor{DRed}{\(\checkmark\) DECISIVE EVIDENCE ALREADY RETURNED}}\\[-1pt]
{\casecartoon\scriptsize\textcolor{DRed}{\(\times\) EIGHT MORE SEARCHES STILL FOLLOW}}\par
\vspace{2pt}
\textbf{[Calls 2--4]} queries: ``\textquotedbl{}Stray\textquotedbl{} game Nintendo Switch'',\par
\hspace*{1em}``\textquotedbl{}Stray\textquotedbl{} Nintendo Switch release date'',\par
\hspace*{1em}``\textquotedbl{}Stray\textquotedbl{} game release date Nintendo Switch''\par
\textbf{[Calls 5--8]} queries: ``\textquotedbl{}Stray\textquotedbl{} game announcement Nintendo Switch'',\par
\hspace*{1em}``\textquotedbl{}Stray\textquotedbl{} Nintendo Switch launch'', ``\textquotedbl{}Stray\textquotedbl{} game release schedule'',\par
\hspace*{1em}``\textquotedbl{}Stray\textquotedbl{} developer interview''\par
\textbf{[Call 9]} text\_search\_tool(query=``\textquotedbl{}Stray\textquotedbl{} Nintendo Switch'')\par
\vfill
\textbf{[Answer]} November 19, 2024 \quad \textcolor{DRed}{[correct, but 9 calls]}\par
}
\end{minipage}}
\hfill
\fcolorbox{DGreen}{CleanRow}{%
\begin{minipage}[t][2.55in][t]{0.455\textwidth}
\begin{minipage}[c]{0.69\linewidth}
{\casecartoon\large\textcolor{DGreen}{PATH B: Full \methodname{} + Reg.}}\\[-1pt]
{\casecartoon\scriptsize\textcolor{black!62}{CORRECT OUTCOME, STOP AFTER EVIDENCE}}
\end{minipage}\hfill
\fcolorbox{DGreen}{white}{%
\begin{minipage}[c]{0.22\linewidth}
\centering
{\casecartoon\large\textcolor{DGreen}{1 CALL}}\\[-2pt]
{\casecartoon\scriptsize answer judge = 1.0}
\end{minipage}}

\vspace{3pt}\hrule\vspace{5pt}
{\ttfamily\fontsize{7.2}{8.8}\selectfont
\textbf{[Call 1]} text\_search\_tool(\par
\hspace*{1em}query=``Stray game release date Nintendo Switch'')\par
\textbf{[Return 1]} Summary: \ldots{} \emph{Stray} is set to release on Nintendo Switch on \textbf{November 19, 2024}.\par
\vspace{6pt}
{\casecartoon\small\textcolor{DGreen}{GOAL ALIGNED}}\\[-1pt]
{\caseround\small Release-date evidence requested}\\[5pt]
{\casecartoon\small\textcolor{DGreen}{EVIDENCE ACQUIRED}}\\[-1pt]
{\caseround\small Required date extracted from Return 1}\\[5pt]
{\casecartoon\small\textcolor{DGreen}{STOP}}\\[-1pt]
{\caseround\small Evidence goal satisfied; no repeated call}
\vfill
\textbf{[Answer]} \textcolor{DGreen}{November 19, 2024} \quad \textcolor{DGreen}{[correct, then stop]}\par
}
\end{minipage}}

\vspace{5pt}
\fcolorbox{black!78}{yellow!18}{%
\begin{minipage}{0.965\textwidth}
\begin{minipage}[c]{0.24\linewidth}
\centering
{\casecartoon\scriptsize\textcolor{black!65}{SAME CORRECT ANSWER}}\\[-1pt]
{\casecartoon\LARGE\textcolor{DRed}{9}\ \(\boldsymbol{\rightarrow}\)\
\textcolor{DGreen}{1}\ {\large CALLS}}
\end{minipage}
\hfill
\begin{minipage}[c]{0.70\linewidth}
{\casecartoon\large\textcolor{blue!60!black}{88.9\% FEWER TOOL CALLS}}\\[-1pt]
{\caseround\small Full \methodname{} learns when the necessary evidence is sufficient---not merely which tool can retrieve it.}
\end{minipage}
\end{minipage}}
\endgroup
\caption{Real MMSearch case \texttt{mmsearch\_end2end\_only\_image-64}. The image and question come from the dataset record; displayed calls, returns, answers, and answer-judge outcomes come from saved rollouts, with long returns excerpted. Under the same training pool and three-tool interface, both policies answer correctly, but Full \methodname{} + Reg. stops after the first decisive result instead of repeating a satisfied goal.}
\label{fig:app_case_study}
\end{figure*}

Figure~\ref{fig:app_case_study} gives a same-pool, same-interface diagnostic comparison on one real MMSearch example using abridged observable trajectory fields from saved rollout JSONL files.
The goal-only policy receives the answer-critical Nintendo Switch release date on its first call but continues through nine text searches that repeatedly target the same release-date goal.
Full \methodname{} with the non-repeated-goal regularizer retrieves the release date in one call and then answers correctly.
The example captures the central behavioral insight behind \methodname{}: efficient tool use requires learning which returned evidence is necessary, not merely learning to invoke an appropriate tool.

\section{Additional Experiments}
\label{app:additional_experiments}

\subsection{Experimental Setup}
\label{app:experimental_setup}

Table~\ref{tab:app_experimental_setup} consolidates the implementation and evaluation settings for the current search-only \modelname{} OOD-evaluation protocol.
The training interface deliberately excludes crop/zoom, whereas evaluation exposes the common three-tool interface; this separation is what makes crop/zoom an unseen tool capability rather than a trained behavior.
Unless a subsection states otherwise, additional evaluations use the same selected checkpoint, benchmark examples, tool schemas, and decoding configuration as the corresponding main-paper comparison.
The reward-composition diagnostics in the next subsection retain their original training pools, which are stated alongside the corresponding rows.
The visual toolset-transfer subsection is an explicit exception: it uses the
frozen step-180 checkpoint so that the two-tool and three-tool conditions are
paired evaluations of identical weights.

\begin{table}[t]
\centering
\small
\setlength{\tabcolsep}{3.5pt}
\renewcommand{\arraystretch}{1.06}
\begin{tabular}{@{}p{0.31\columnwidth}p{0.61\columnwidth}@{}}
\toprule
Item & Setting \\
\midrule
Base policy & Qwen3-VL-8B-Instruct; vision tower frozen \\
Training pool & 4,855 search-oriented, no-zoom examples \\
Training tools & \toolcall{image\_search\_tool}, \toolcall{text\_search\_tool} \\
Evaluation tools & \toolcall{image\_zoom\_in\_tool}, \toolcall{image\_search\_tool}, \toolcall{text\_search\_tool} \\
GRPO sampling & Group size $G=8$; training batch size 128 \\
Sequence limits & 32K context; 16K response; at most 10 interaction turns \\
Optimization & AdamW; learning rate $1\!\times\!10^{-5}$; 20 epochs \\
Evaluation size & 2,317 examples for the matched main protocol; 4,417 for full-split analyses \\
Answer scoring & Shared VLM judge for open-ended search; exact match for visual multiple choice \\
NTEP accounting & Info hit $+0.7$; goal hit $+0.3$; goal miss $-0.3$; duplicate goal $-0.1$; scale $0.5$ \\
\bottomrule
\end{tabular}
\caption{Implementation and evaluation setup for the current search-only \modelname{} OOD-evaluation protocol. Crop/zoom is withheld during training and introduced only through the three-tool evaluation interface.}
\label{tab:app_experimental_setup}
\end{table}

\subsection{Exact Reward-Ablation Values}

Figure~\ref{fig:ablation_components} summarizes the component ablation in the main text; Table~\ref{tab:app_ablation_values} reports the corresponding three-decimal values.
Full \methodname{} and the goal-only process-reward row use the same historical 7,774-example training pool.
Answer Reward Only is retained as a 3,859-example legacy diagnostic, matching the note in Section~4.4, and serves as a call-efficiency reference point.

\begin{table}[t]
\centering
\small
\setlength{\tabcolsep}{4.0pt}
\resizebox{\columnwidth}{!}{%
\begin{tabular}{@{}lrrrrrrr@{}}
\toprule
Setting & Search Avg. & Visual Avg. & Overall Avg. & Crop & Image & Text & Total \\
\midrule
Full \methodname{} & 59.438 & 83.359 & 69.690 & 0.518 & 0.407 & 0.623 & 1.549 \\
w/o Info. Acq. \& Non-Rep. Goal Reg. & 34.580 & 83.374 & 55.492 & 0.511 & 3.178 & 1.671 & 5.360 \\
Answer Reward Only & 58.016 & 85.124 & 69.634 & 1.561 & 0.718 & 0.828 & 3.107 \\
\bottomrule
\end{tabular}}
\caption{Reward-component diagnostics (values shown to three decimals). Accuracy columns are split-macro percentages and tool columns are average calls per example; totals are computed from unrounded per-tool counts and can differ by $0.001$ from the visible component sum. The first two rows share the 7,774-example historical pool; Answer Reward Only uses a 3,859-example legacy pool and is shown as a diagnostic rather than a data-matched ablation.}
\label{tab:app_ablation_values}
\end{table}

\subsection{Inference-Budget Sweep}
\label{app:budget_sweep}

Table~\ref{tab:app_budget_sweep} reports the inference-budget sweep referenced in Appendix~\ref{app:theory}: the same Full \methodname{} checkpoint is evaluated on the full 4,417-example suite with the maximum number of assistant turns varied over $B\in\{5,10,15,20\}$.
Accuracy and realized tool use are essentially flat across budgets, confirming that the policy's demand for calls is set by the evidence path rather than by the available budget.

\begin{table}[t]
\centering
\small
\setlength{\tabcolsep}{6.0pt}
\renewcommand{\arraystretch}{1.05}
\resizebox{\columnwidth}{!}{%
\begin{tabular}{@{}crrrr@{}}
\toprule
Budget $B$ & Search Avg. & Visual Avg. & Overall Avg. & Total calls \\
\midrule
5 & 59.256 & 83.051 & 69.454 & 1.548 \\
10 & 59.438 & 83.359 & 69.690 & 1.549 \\
15 & 59.047 & 83.851 & 69.677 & 1.545 \\
20 & 58.700 & 83.926 & 69.512 & 1.548 \\
\bottomrule
\end{tabular}}
\caption{Inference-budget sweep for Full \methodname{} on the full 4,417-example suite, using the same checkpoint and scoring as Table~\ref{tab:app_ablation_values} ($B=10$ is the main protocol). Accuracy columns are split-macro percentages; Total calls is the average number of tool calls per example.}
\label{tab:app_budget_sweep}
\end{table}

\subsection{Tool-Need Counterfactual}

Table~\ref{tab:tool_necessity} tests whether the trained checkpoint actually needs external tools.
Both modes use the same \modelname{} weights and the same examples.
The end-to-end mode removes the tool interface, while the agentic mode enables crop/zoom, image search, and text search.
The counterfactual shows that tools change many otherwise wrong predictions into correct ones, especially on search-oriented benchmarks, while introducing relatively few cases where a correct direct answer becomes wrong.

\begin{table}[t]
\centering
\small
\setlength{\tabcolsep}{3.0pt}
\resizebox{\columnwidth}{!}{%
\begin{tabular}{@{}lrrrrrrrrr@{}}
\toprule
Benchmark & $N$ & E2E Acc. & Agent Acc. & E2E$\times$,Agent$\checkmark$ & E2E$\checkmark$,Agent$\times$ & Crop & Image & Text & Total \\
\midrule
MMSearch & 171 & 14.6 & 63.7 & 86 (50.3) & 2 (1.2) & 0.07 & 0.73 & 0.90 & 1.70 \\
HR-MMS & 305 & 4.3 & 33.4 & 94 (30.8) & 5 (1.6) & 0.29 & 0.48 & 1.22 & 1.98 \\
InfoSeek & 500 & 25.6 & 56.8 & 173 (34.6) & 17 (3.4) & 0.05 & 0.98 & 0.82 & 1.85 \\
MAT & 150 & 61.3 & 84.7 & 38 (25.3) & 3 (2.0) & 0.03 & 0.62 & 1.37 & 2.02 \\
V* & 191 & 86.9 & 90.6 & 11 (5.8) & 4 (2.1) & 1.07 & 0.00 & 0.00 & 1.07 \\
HR-4K & 500 & 79.4 & 81.2 & 44 (8.8) & 35 (7.0) & 1.08 & 0.02 & 0.01 & 1.10 \\
HR-8K & 500 & 75.8 & 78.4 & 62 (12.4) & 49 (9.8) & 1.06 & 0.01 & 0.01 & 1.08 \\
\midrule
\textbf{All} & 2317 & 51.8 & 68.8 & 508 (21.9) & 115 (5.0) & 0.61 & 0.38 & 0.50 & 1.48 \\
\bottomrule
\end{tabular}}
\caption{Same-checkpoint tool-necessity counterfactual. E2E uses no tools; Agent enables \toolcall{image\_zoom\_in\_tool}, \toolcall{image\_search\_tool}, and \toolcall{text\_search\_tool}. Parentheses report percentages of the benchmark split. Per-tool and total call columns are averaged independently from unrounded counts, so their displayed sums can differ by $0.01$.}
\label{tab:tool_necessity}
\end{table}

\subsection{Trajectory-Label Distribution}

For the behavioral analysis, we label 300 tool-using trajectories per model, stratified across the seven benchmarks.
Labels are assigned per call with the deterministic priority redundant \(>\) off-goal \(>\) wrong tool \(>\) evidence returned but unused \(>\) necessary-and-used.
Table~\ref{tab:app_taxonomy_values} reports the exact distribution behind Figure~\ref{fig:failure_modes} and the trajectory analysis in Section~4.4.

\begin{table}[t]
\centering
\small
\setlength{\tabcolsep}{4.0pt}
\resizebox{\columnwidth}{!}{%
\begin{tabular}{@{}lrrrrrrr@{}}
\toprule
Model & Calls/traj. & Off-goal & Wrong tool & Redundant & Unused evidence & Necessary-used & Case-5 \\
\midrule
Full \methodname{} & 1.59 & 1.9 & 14.5 & 1.0 & 3.8 & 78.8 & 58.7 \\
\methodname{} w/o Reg. & 2.35 & 4.5 & 17.9 & 13.5 & 3.8 & 60.2 & 29.3 \\
SenseNova-MARS-8B & 2.24 & 1.8 & 18.2 & 8.8 & 2.2 & 69.0 & 39.0 \\
Zero-shot Qwen3-VL-8B & 2.16 & 4.2 & 24.6 & 4.9 & 2.5 & 63.8 & 50.0 \\
\bottomrule
\end{tabular}}
\caption{Exact trajectory-label distribution on 300 judge-labeled tool-using trajectories per model. Failure-mode columns are percentages of tool calls; Case-5 is the percentage of trajectories where all calls are necessary-used and the final answer is correct.}
\label{tab:app_taxonomy_values}
\end{table}
\begin{table}[ht!]
\centering
\small
\setlength{\tabcolsep}{4pt}
\renewcommand{\arraystretch}{1.06}
\resizebox{\columnwidth}{!}{%
\begin{tabular}{@{}lrrrrr@{}}
\toprule
& & \multicolumn{2}{c}{Accuracy (\%)} & \multicolumn{2}{c}{Avg.\ calls} \\
\cmidrule(lr){3-4}\cmidrule(lr){5-6}
Benchmark & $N$ & Search tools & $+$Zoom & Search tools & $+$Zoom \\
\midrule
V* Bench & 191 & 75.39 & \textbf{85.34} & 1.707 & \textbf{1.230} \\
HR-Bench 4K & 800 & 77.12 & \textbf{83.50} & 1.651 & \textbf{1.318} \\
HR-Bench 8K & 800 & 72.50 & \textbf{78.12} & 1.679 & \textbf{1.334} \\
\midrule
\textbf{Average} & --- & 75.01 & \textbf{82.32} & 1.679 & \textbf{1.294} \\
\bottomrule
\end{tabular}}
\caption{Exact values for the visual OOD toolset transfer in Figure~\ref{fig:ood_visual_toolset}. Both conditions use the frozen step-180 checkpoint and identical decoding; only the exposed tool interface differs. Averages are computed from unrounded values.}
\label{tab:app_toolset_values}
\end{table}
\subsection{Toolset-Transfer Exact Values}
\label{app:toolset_values}

Table~\ref{tab:app_toolset_values} reports the exact per-benchmark values behind Figure~\ref{fig:ood_visual_toolset}.
Accuracy improves and average calls drop on every visual split when the previously unseen zoom tool is added; the largest accuracy gain ($+9.95$ points on V*) coincides with the largest call reduction.

\begin{figure*}[t]
\centering
\includegraphics[width=0.92\textwidth]{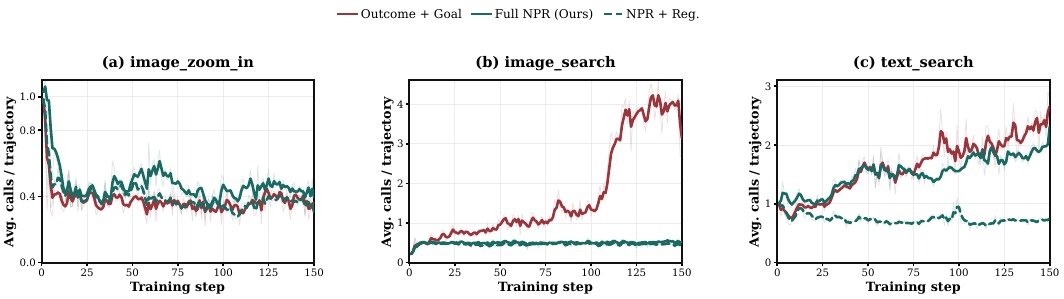}
\caption{Per-tool training dynamics (average per-trajectory invocation count). Zoom converges to the same level under all three configurations, so necessary visual operations are preserved; saturation lands primarily on image search under goal-only supervision; text search is preserved by information supervision and compressed only by the non-repeated-goal regularizer.}
\label{fig:app_tool_dynamics}
\end{figure*}

\begin{figure}[ht!]
\centering
\includegraphics[width=\columnwidth]{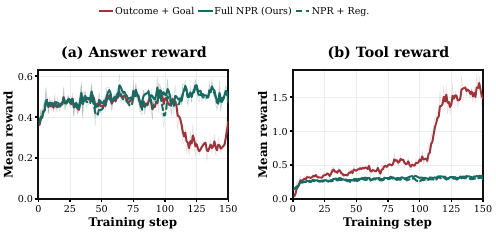}
\caption{Reward dynamics over GRPO training on the historical pool. (a) Answer reward: the goal-only configuration collapses after step ${\sim}100$, while both information-supervised configurations hold near $0.5$. (b) Process reward: the goal-only configuration inflates its tool reward to ${\approx}1.6$ by crediting calls whose returns are never used; the information-supervised configurations remain flat. The simultaneous process-reward spike and answer-reward collapse is the training-time signature of saturation-style tool calling.}
\label{fig:app_reward_dynamics}
\end{figure}

\subsection{Training Dynamics}
\label{app:training_dynamics}

Figures~\ref{fig:app_reward_dynamics} and~\ref{fig:app_tool_dynamics} track the archived GRPO process-reward runs on the historical 7,774-example pool. Plot legends retain their historical run names for reproducibility; \emph{Outcome+Goal} corresponds to the goal-only row in Table~\ref{tab:app_ablation_values}, and \emph{NPR + Reg.} corresponds to Full \methodname{}.
Three observations support the reward design.
First, pre-call alignment alone is unstable: the goal-only run inflates its process reward late in training while its answer reward collapses---the training-time signature of saturation-style calling---whereas both information-supervised runs hold a stable answer reward throughout.
Second, the effect of \methodname{} is selective rather than uniform: the per-trajectory rate of \toolcall{image\_zoom\_in\_tool} converges to the same level under all three configurations, so necessary visual operations are preserved by every variant, while the goal-only run's \toolcall{image\_search\_tool} rate climbs toward four calls per trajectory and the information-supervised runs stay near $0.5$.
Third, information supervision does not suppress legitimate retrieval: on \toolcall{text\_search\_tool} the w/o-regularizer run tracks the goal-only run closely, and only the non-repeated-goal regularizer compresses it further toward the efficient operating point reported in the main results.

\subsection{Unified Evaluation Protocol and Baseline Adapter}
\label{app:eval_protocol}

All rows of Table~\ref{tab:main_results} are produced by one evaluation harness that fixes the datasets, prompts, tool schemas, tool backends, image budget, interaction budget, decoding budget, and scoring protocol across systems.
Policies decode with temperature $1.0$, top-$p$ $1.0$, top-$k$ $20$, presence penalty $1.5$, and a fixed seed, under a 32K context, a 16K generated-token cap, at most 10 interaction turns, and a shared maximum image budget of $8{,}294{,}400$ pixels.
Visual multiple-choice benchmarks are scored by a single deterministic option-matching rule shared by every row.
Open-ended search answers are scored by exact match first and otherwise by a vision-capable judge (Qwen3-VL-Plus, temperature $0$) that receives the image, question, ground truth, and prediction under the same rubric as the training-time judge; empty predictions are scored $0$ without judge involvement.

External RL-agent checkpoints are evaluated through a generic in-process multi-turn driver: it parses each model's native tool-call dialect, dispatches every call to the same three tool backends used by \modelname{} (normalized-coordinate crop with native smart-resize, reverse image search, and top-3 summarized text search), and feeds the returns back in the model's expected format.
Each checkpoint required only minimal, additive wiring---registering the missing tools inside its own agent loop---so these rows measure the transfer of released weights to the common interface under identical evidence sources and budgets.
For closed-source APIs, parameters a provider does not accept are left at provider defaults and recorded per row, and providers whose serving stack rejects the native tool schema receive the same tools through text-format tool-call parsing.

\subsection{Limitations}
\label{app:limitations}

We note the scope of the current study.
First, path quality is bounded by the teacher: NTEPs are distilled by a stronger model, and the completion branch, which reconstructs paths for failed rollouts, is intrinsically harder than pruning successful ones; the retained outcome reward anchors training against residual path noise, but the dependence remains.
Second, step-wise semantic judging adds serving overhead relative to outcome-only RL; bounded judge concurrency and response caps keep this manageable at our scale, and the judge is needed only during training.
Third, our experiments cover three tools and predominantly English retrieval; because the NTEP interface is tool-agnostic (Section~3), richer tool families and multilingual corpora are natural extensions rather than redesigns.
Fourth, the non-repeated-goal regularizer selects an efficiency-leaning operating point on the accuracy--efficiency frontier; applications that value maximal retries can tune the duplicate penalty accordingly.

\begin{table}[ht!]
\centering
\small
\setlength{\tabcolsep}{4.0pt}
\renewcommand{\arraystretch}{1.08}
\begin{tabular}{@{}p{0.42\columnwidth}p{0.46\columnwidth}@{}}
\toprule
Item & Value \\
\midrule
Audit size & 100 judge-involved cases \\
Scope & Search-answer scoring; trajectory-labeling analysis \\
Benchmarks & Search-oriented splits and trajectory-analysis samples \\
Blinding & Model identity and judge score hidden \\
Inputs shown & Image, question, model answer, tool transcript, returned evidence when applicable \\
Labels & Correctness; necessary/unnecessary use; non-necessary failure mode \\
Agreement & 97.0\% (97/100) \\
Cohen's $\kappa$ & 0.936 \\
Use of audit & Reliability check only; not used for selection or tuning \\
\bottomrule
\end{tabular}
\caption{Blinded human audit of judge reliability across judge-involved settings.}
\label{tab:app_human_audit}
\end{table}

\section{Human Audit}
\label{app:human_audit}

The human audit validates the automated judgments used for open-ended search-answer scoring and trajectory-level behavior analysis.
We sample 100 judge-scored cases from the judge-involved settings and hide both model identity and judge decision from the annotator.
Each audit row contains the image, question, model answer, tool transcript, and returned evidence when applicable.
Annotators assign the same label types used by the automated judge, including answer correctness, necessary-versus-unnecessary tool use, and the failure category for non-necessary calls.
Table~\ref{tab:app_human_audit} summarizes the audit protocol and agreement.

Agreement is uniform across systems: $98\%$ on \modelname{} cases and $96\%$ on the strongest baseline's cases, with one judge false positive and two false negatives among the 100 audited decisions.
All three disagreements are boundary judgments---synonym granularity (e.g., ``contemporary'' vs.\ ``modern'') and a magnitude misreading---rather than systematic bias toward any model.

This audit is not used for model selection, reward tuning, or post-hoc correction of reported results.
The visual multiple-choice benchmarks do not use automated judging for correctness; they are scored by answer extraction and exact matching.

\section{Non-Commercial Use Statement}

This work uses MMSearch (end-to-end, image-only), InfoSeek, V*Bench, HR-Bench 4K, and HR-Bench 8K. Although the code or annotations of these benchmarks are released under permissive licenses such as MIT or Apache-2.0, some of their underlying images and source data remain subject to non-commercial or research-only terms imposed by the original data providers. The authors confirm that these benchmarks and their underlying data were used solely for non-commercial academic research and were not used for any commercial activity, in accordance with the applicable upstream licenses and terms of use.

\end{document}